\documentclass{article}

\usepackage[final, nonatbib]{nips_2018}

\usepackage[utf8]{inputenc} 
\usepackage[T1]{fontenc}    
\usepackage{url}            
\usepackage{booktabs}       
\usepackage{amsfonts}       
\usepackage{nicefrac}       
\usepackage{microtype}      
\usepackage[numbers]{natbib}
\usepackage{graphicx}
\usepackage{amsmath}

\usepackage[breaklinks]{hyperref}
\usepackage{breakurl}

\title{Honey Authentication with Machine Learning Augmented Bright-Field Microscopy}

\author{
  Peter He \\
  Department of Computing\\
  Imperial College London\\
  London, United Kingdom \\
  \texttt{ph1718@ic.ac.uk} \\
   \And
  Alexis Gkantiragas \\
  Department of Molecular Biology \\
  University College London \\
  London, United Kingdom \\
  \texttt{alexis.gkantiragas.17@ucl.ac.uk} \\
  \And
  Gerard Glowacki \\
  Hampton Court House \\
  London, United Kingdom \\
  \texttt{gerag7@hchschool.com} \\
}

\begin{document}

\maketitle

\begin{abstract}
Honey has been collected and used by humankind as both a food and medicine for thousands of years. However, in the modern economy, honey has become subject to mislabelling and adulteration making it the third most faked food product in the world. The international scale of fraudulent honey has had both economic and environmental ramifications. In this paper, we propose a novel method of identifying fraudulent honey using machine learning augmented microscopy.

\end{abstract}

\section{Introduction}

\subsection{Background}

Honey is the natural sweet substance produced by honey bees from the nectar of plants or from secretions of living parts of plants or excretions of plant sucking insects on the living parts of plants which bees collect and transform by combining with specific substances of their own deposit, dehydrate, store and leave in the honey comb to ripen and mature \cite{codex}. It consists of various sugars as well as organic acids, enzymes and solid particles (such as pollen) derived from honey collection. The flavour and aroma of honey varies, but is derived from the honey's plant origin. \cite{codex} For millennia, humankind has collected and used honey for both culinary and medicinal purposes \cite{bell_2007}. In modern times, honey production is a global industry and its bees are essential for pollination and agriculture in the developed world \cite{presentation}. Furthermore initiatives such as Bees for Development have sprung up aiming to use beekeeping as a means to alleviate poverty in the developing world \cite{bfd}.

Honey, however, is also the third most faked food product globally and is often subject to mislabelling (selling one honey as another) and adulteration (dilution of honey using sugar syrup, for instance) \cite{moore_spink_lipp_2012, chen_xue_ye_zhou_chen_zhao_2011}. This is especially prevalent with more expensive regional honeys such as acacia and manuka honey, for which global consumption is allegedly greater than global production \cite{usborne_2014}. The truly global nature of this food fraud was most notably illuminated by the so-called 'Honeygate' scandal in which it was discovered that  Chinese honey was being transshipped through Germany (and thereby labelled as German honey) and imported to the USA by a number of food suppliers \cite{leeder_2011, swaine_2013, trinidad}.

Such fraudulent practices have a negative impact on producers of genuine honey products as production costs for the fake products are significantly lower. This leads to lower profit margins for genuine producers as the market price for their produce falls sometimes forcing them to leave the market and discouraging new players from entering. It also means that poorer regions producing valuable regional honeys lose out on potential revenue. Furthermore, beekeeping practices in the production of fraudulent honey products are often very much sub-par when compared to those producing genuine honey. These can include the over-harvesting of honey which can adversely affect the bee colonies involved and malnutrition of the colony \cite{cairns_villanueva-gutierrez_koptur_bray_2005}.

A robust and scalable method of honey authentication would thus economically incentivise the production of genuine honey products, bring forth many international development opportunities and potentially spark a rise in amateur beekeeping in urban areas (which would indeed aid urban bee populations).

\subsection{Existing Methods of Honey Authentication}

Due to pervasiveness of fraudulent honey products, the authentication of honey has become an active area of research with nations such as New Zealand, seeking to protect their valuable Manuka honey exports and the European Union trying to protect domestic consumer and producer interests \cite{criteria, honey_2015}.

Current honey authentication procedures include quantitative polymerase chain reaction (qPCR) \cite{criteria}, nuclear magnetic resonance spectroscopy (NMR) \cite{beretta_caneva_regazzoni_bakhtyari_facino_2008}, liquid chromatography mass spectrometry (LC-MS) \cite{criteria}, near-infrared spectroscopy (NIR) \cite{chen_xue_ye_zhou_chen_zhao_2011}, and microscopy (as different honeys contain visually distinct pollen from different floral sources) \cite{criteria, sniderman_matley_haberle_cantrill_2018}. Other tests exist for the identification of specific honeys such as testing for methylglyoxal and leptosperin in manuka honey \cite{oelschlaegel_gruner_wang_boettcher_koelling-speer_speer_2012, kato_fujinaka_ishisaka_nitta_kitamoto_takimoto_2014} though these have proven insufficient due to the fact that methylglyoxal is unstable and its levels change over time \cite{criteria}. All such procedures are conducted in laboratories by specialists and the analytical tests (qPCR, NIR, NMR and LC-MS) require specialised equipment which can prove expensive. The authentication of honey through microscopy proves difficult too due to the pollen-identifying expertise required of the operator \cite{criteria, soares_amaral_oliveira_mafra_2017}.

The current state-of-the-art in manuka honey authentication makes use of four chemical markers (2'-methoxyacetophenone, 2-methoxybenzoic acid, 4-hydroxyphenyllactic acid and 3-phenyllactic acid) and a test for the DNA of manuka pollen (qPCR) \cite{criteria}. The method is, however, only applicable to authenticating manuka honey.

\section{Method and Experiments}
\subsection{Overview}
We propose a method of honey authentication which utilises microscopy while at the same time eliminating the need for an expert operator. This is achieved through the automation of pollen classification using machine learning techniques. In this way, honey authentication via pollen can be carried out by non-specialists and at scale (in contrast to centralised testing facilities). Furthermore, our method lays the groundwork for accelerating more advanced quantitative approaches to honey authentication using pollen. 

Our proposed pipeline comprises of two separate parts: the pollen identification network and the authentication network. Given an image obtained from a microscope, the pollen identification network segments and identifies the botanical origin, density and distribution of the extracted pollen grains. The outputs from the pollen identification network are then passed (alongside any other test results both physical and chemical) into the authentication network which outputs a decision as to whether or not the honey is genuine. This modular pipeline allows the authentication network to be retrained for any purpose while the pollen identification network remains static. Furthermore, it allows the authentication network to be replaced entirely with any other classifier (such as a decision tree, for added interpretability).

\subsection{Data Collection}
Samples of different honeys (manuka, acacia, 'Lithuanian', 'Black Forest', eucalyptus melliodora and thyme) of equal volume were collected and spread thinly across glass slides. The slides were covered with cover slips and put into a camera-mounted Solomark compound bright-field microscope for analysis at 320x zoom. The microscope was able to be controlled both manually as well as with stepper motors. Images were captured at 1080x1080 pixel resolution (see Figure 1) and, for the purposes of a proof-of-concept, the pollen were annotated and labelled into three distinct overarching classes ("round", "triangular" and "spiky") with rectangular bounding boxes. Approximately 2500 pollen were imaged overall. 

\begin{figure}
  \centering
  \includegraphics[scale=0.39]{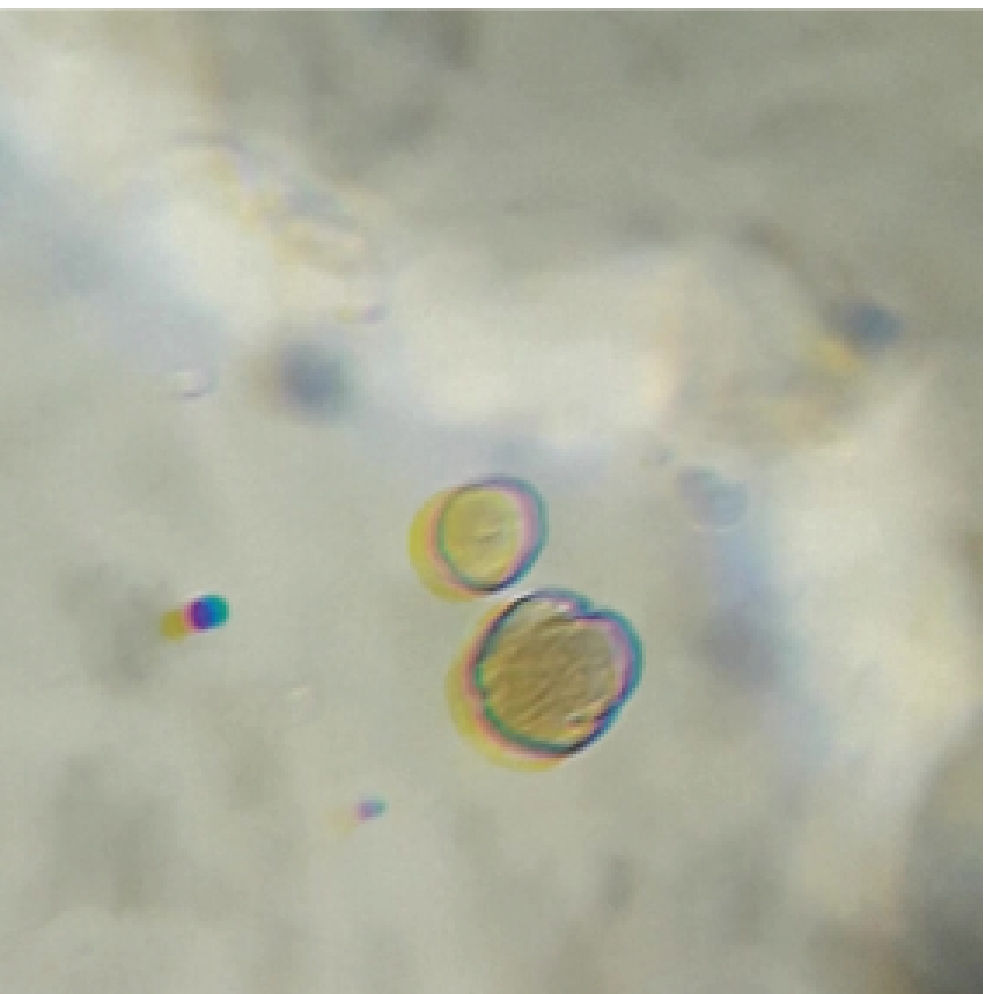} \includegraphics[scale=0.39]{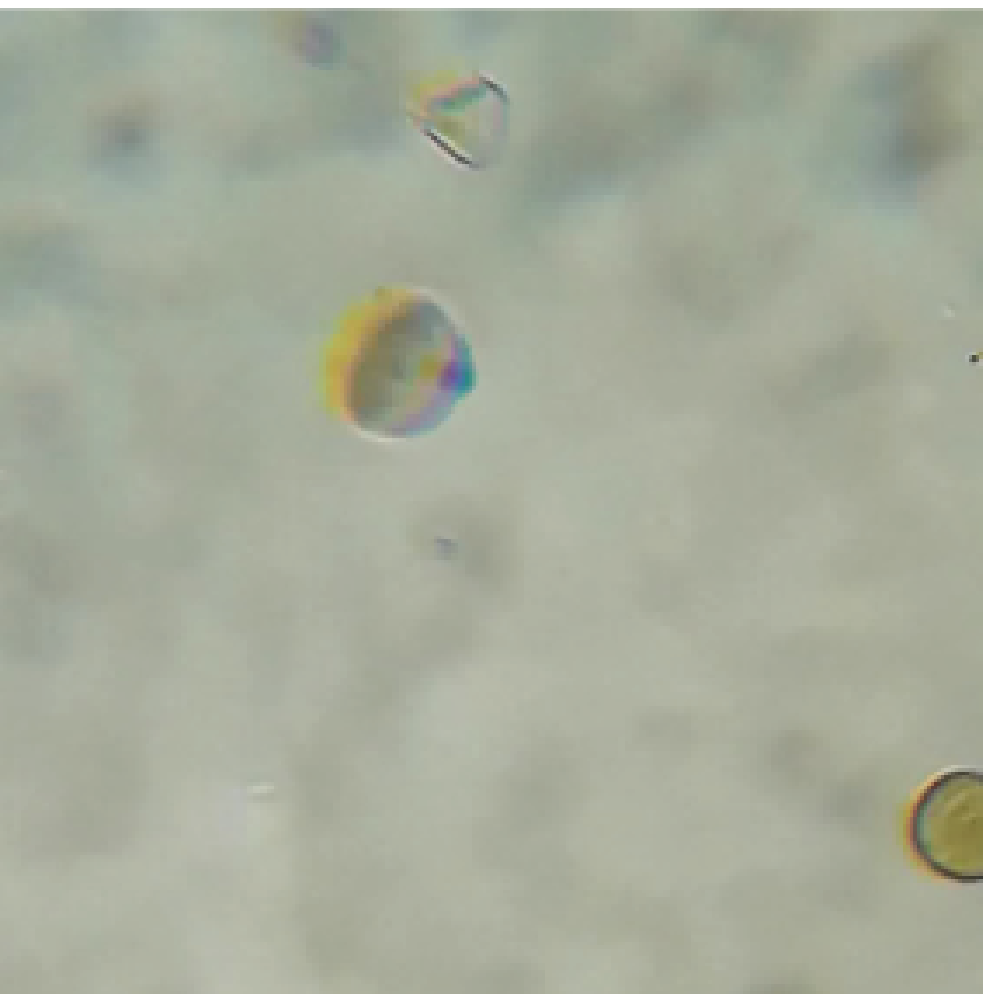} \includegraphics[scale=0.39]{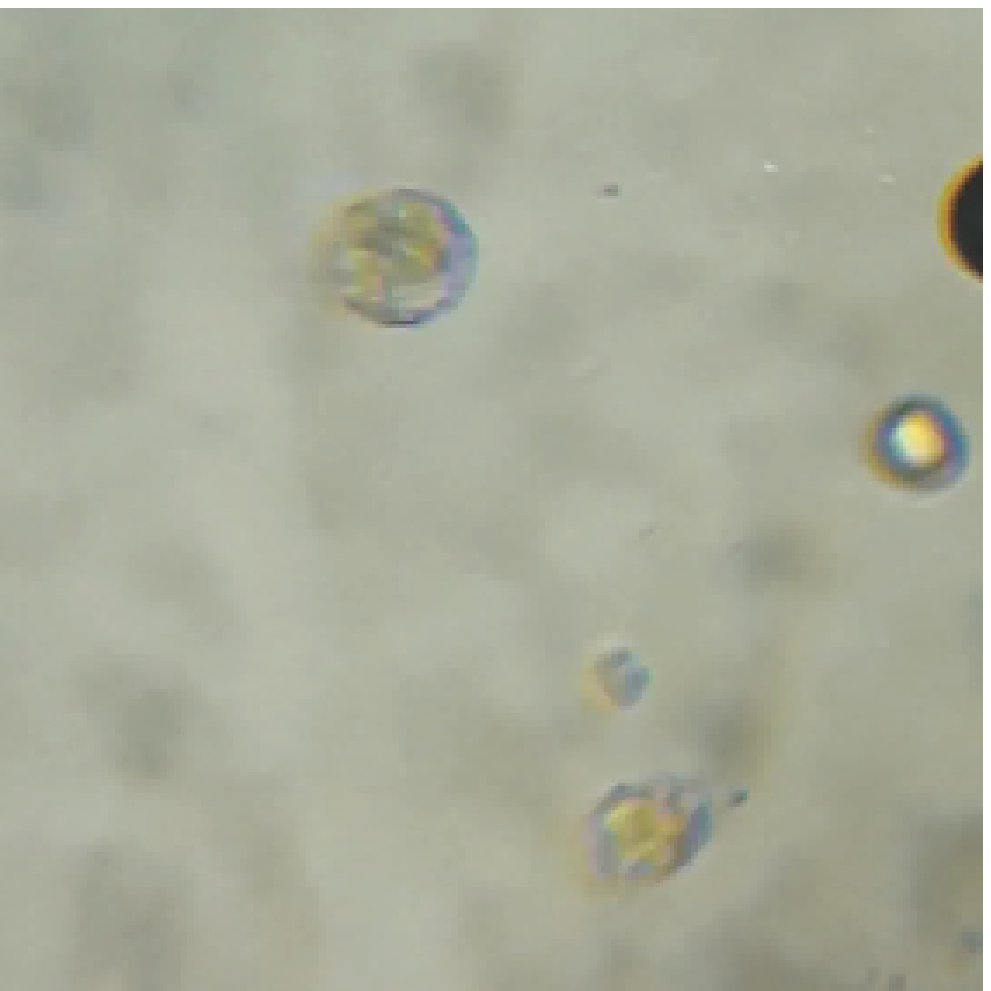} 
  \caption{Bright-field images of pollen from Eucalyptus melliodora honey. The leftmost image is comprised exclusively of "round" pollen while an example of "triangular" pollen can be seen towards the top of the middle image.}
\end{figure}

\begin{figure}
  \centering
  \includegraphics[scale=0.39]{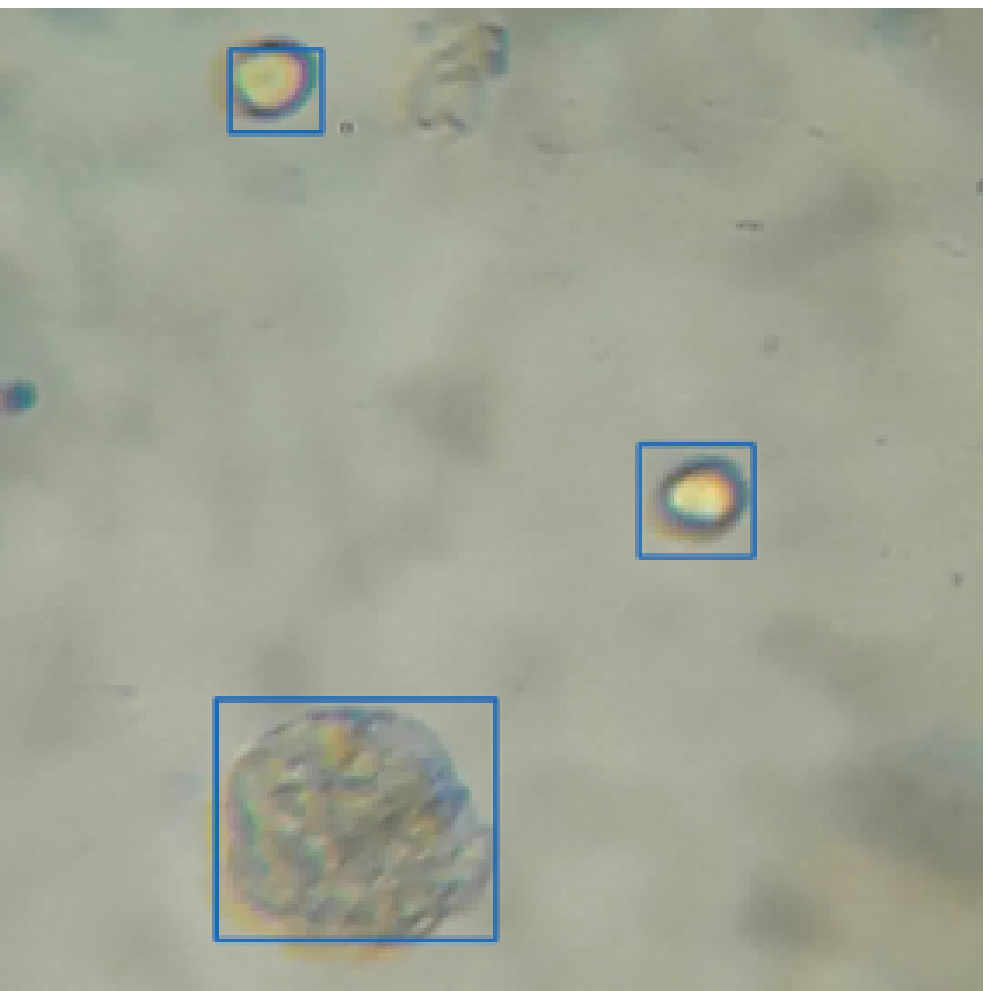} \includegraphics[scale=0.39]{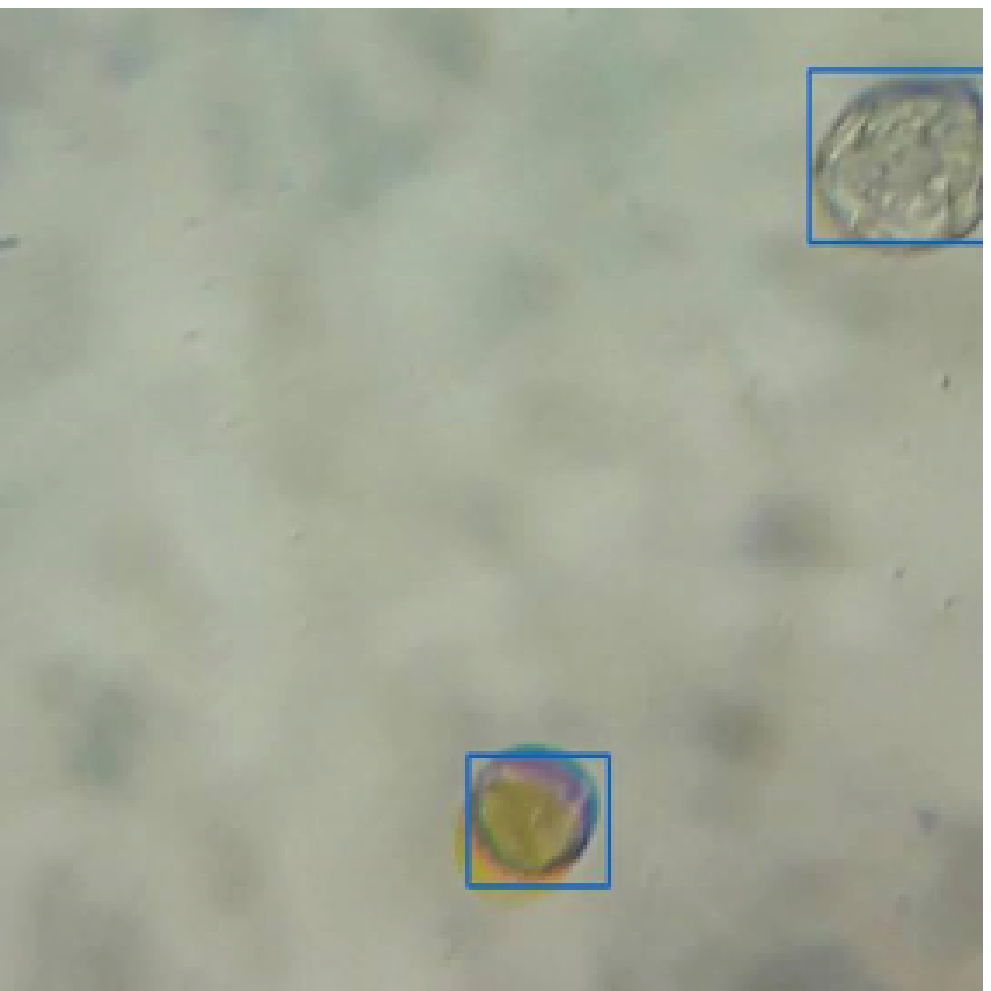} \includegraphics[scale=0.39]{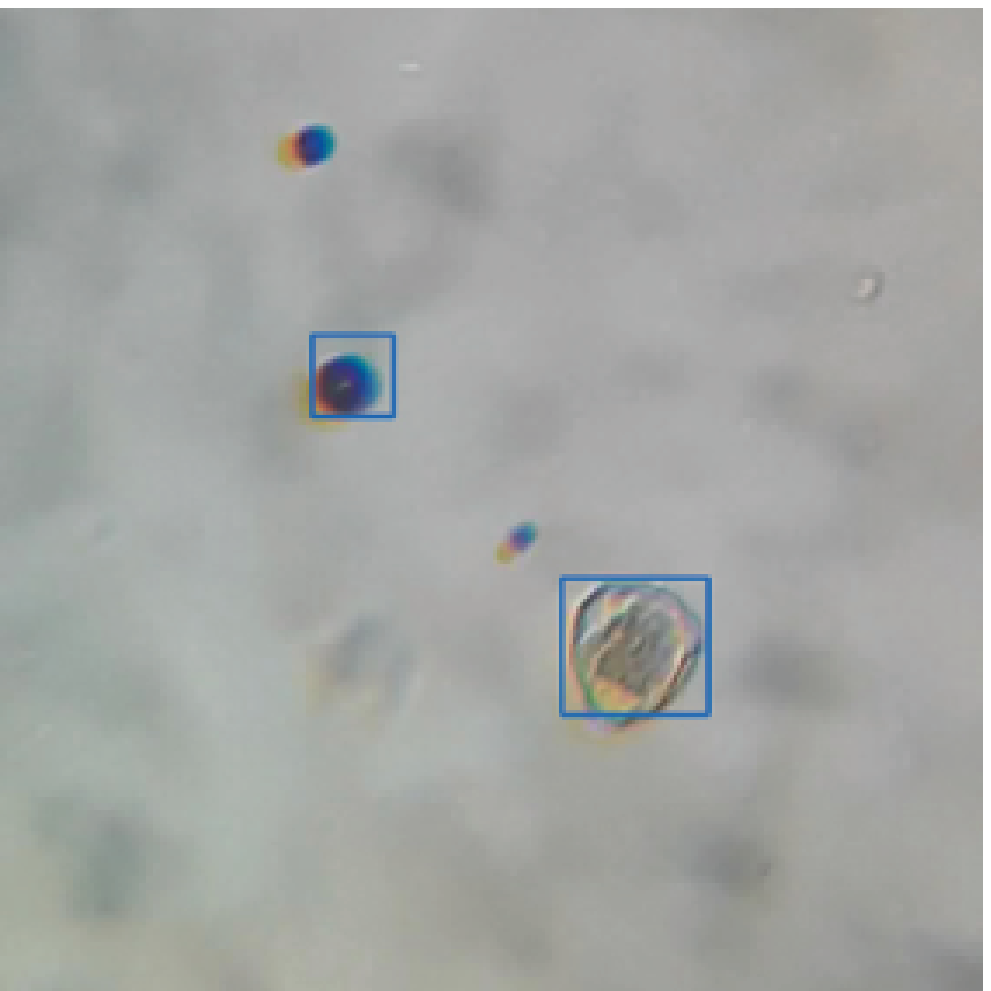} 
  \caption{Bounding boxes over detected pollen. Air bubbles make up the majority of false positives (see the rightmost image) though this can be easily overcome through further training.}
\end{figure}

\subsection{Segmentation Network}
We trained a three-class YOLOv2 network \cite{redmon_farhadi_2017} to detect and segment pollen. We used the standard YOLO loss function \cite{redmon_farhadi_2017} with the architecture displayed in Table \ref{arch}. Results are displayed in Table \ref{metrics} and Figure 2.

\begin{table}[]
\caption{Pollen identification network architecture. * denotes a skip connection which is put through a 64 filter 1 x 1 convolutional layer and reshaped to give a tensor of dimension 13 x 13 x 256.}
\centering
\label{arch}
\begin{tabular}{@{}cccc@{}}
\toprule
\textbf{Type} & \textbf{Filters} & \textbf{Size / Stride} & \textbf{Output}  \\ \midrule
Convolutional & 32               & 3 x 3 / 1              & 416 x 416 x 32   \\
Max Pool      &                  &                        & 208 x 208 x 32   \\
Convolutional & 64               & 3 x 3 / 1              & 208 x 208 x 64   \\
Max Pool      &                  &                        & 104 x 104 x 64   \\
Convolutional & 128              & 3 x 3 / 1              & 104 x 104 x 128  \\
Convolutional & 64               & 1 x 1 / 1              & 104 x 104 x 64   \\
Convolutional & 128              & 3 x 3 / 1              & 104 x 104 x 128  \\
Max Pool      &                  &                        & 52 x 52 x 128    \\
Convolutional & 256              & 3 x 3 / 1              & 52 x 52 x 256    \\
Convolutional & 128              & 1 x 1 / 1              & 52 x 52 x 128    \\
Convolutional & 256              & 3 x 3 / 1              & 52 x 52 x 256    \\
Max Pool      &                  &                        & 26 x 26 x 256    \\
Convolutional & 512              & 3 x 3 / 1              & 26 x 26 x 512    \\
Convolutional & 256              & 1 x 1 / 1              & 26 x 26 x 256    \\
Convolutional & 512              & 3 x 3 / 1              & 26 x 26 x 512    \\
Max Pool*     &                  &                        & 13 x 13 x 512    \\
Convolutional & 1024             & 3 x 3 / 1              & 13 x 13 x 1024   \\
Convolutional & 512              & 1 x 1 / 1              & 13 x 13 x 512    \\
Convolutional & 1024             & 3 x 3 / 1              & 13 x 13 x 1024   \\
Convolutional & 512              & 1 x 1 / 1              & 13 x 13 x 512    \\
Convolutional & 1024             & 3 x 3 / 1              & 13 x 13 x 1024   \\
Convolutional & 1024             & 3 x 3 / 1              & 13 x 13 x 1024   \\
Convolutional & 1024             & 3 x 3 / 1              & 13 x 13 x 1024   \\
Concatenate   &                  &                        & 13 x 13 x 1280   \\
Convolutional & 1024             & 3 x 3 / 1              & 13 x 13 x 1024   \\
Convolutional & 60               & 1 x 1 / 1              & 13 x 13 x 60     \\
Reshape       &                  &                        & 13 x 13 x 10 x 6 \\\bottomrule
\end{tabular}
\end{table}

\begin{table}[]
\caption{Pollen identification network metrics. Results are promising even with limited training time.}
\centering
\label{metrics}
\begin{tabular}{cccc}
\toprule
\textbf{Precision} & \textbf{Sensitivity} & \textbf{Specificity} & \textbf{F1}    \\ \midrule
0.663     & 0.914       & 0.761       & 0.769
\end{tabular}
\end{table}

\subsection{Authentication Network}
We trained a feed-forward neural network with a single hidden layer. The network inputs were pollen counts and overall pollen density (average number of pollen given an arbitrary area). The network was tasked with differentiating five samples of eucalyptus melliodora honey from five samples of manuka honey. All ten samples were classified correctly.

\section{Discussion and Further Work}
It seems, given these preliminary results, that bright-field microscopy would prove a robust and powerful tool for honey authentication when augmented with our machine learning pipeline. Honey samples diluted with sugar syrup can be detected from pollen density analysis and honey samples diluted with cheaper honeys can be detected from pollen distribution comparison. Mislabelled honey samples can be identified through the botanical sources of their pollen. 

The proposed system, however, is unable to identify contamination with heavy metals, pesticides or antibiotics and would thus need to be used in tandem with other chemical tests (which could be integrated into the authentication network). Furthermore, the system would be unable to authenticate ultra-filtered honey samples where pollen is not present though this is sometimes an indicator of adulteration \cite{trinidad}.

In order to bring the proposed system from infancy to maturity, a more significant pollen dataset would need to be gathered. This would allow the dataset to capture the full diversity of pollen in honey \cite{sniderman_matley_haberle_cantrill_2018}. Problems may also arise from different imaging protocols and microscope hardware. It would thus be useful to explore possible generalization methods for robust pollen representations beyond regularisation such as the adverserial training seen in DeepMedic \cite{kamnitsas_baumgartner_ledig_newcombe_simpson_kane_menon_nori_criminisi_rueckert_et}.

The impacts of the introduction of our system to the honey industry and regulators would likely include lower capital and labour costs for honey testing and, as a result of this, the development of more decentralised honey authentication. Nevertheless, in order for any form of honey authentication to be effective, robust regulatory frameworks must also be put in place and this, of course, very much relies on governments and policy-makers.

\section{Conclusions}
We have proposed a new spin to an old approach to honey authentication capable of detecting diluted and mislabelled honey through quantitative means and at a lower labour and capital cost. We have presented a proof-of-concept for the proposed solution with promising results forming a strong case for further investigation using more state-of-the-art techniques. Our system makes use of a modular pipeline with a separate final authentication network and can therefore be integrated with existing authentication processes.

If developed further, our system will likely prove a powerful tool in the fight against fraudulent honey, an industry which has cost livelihoods, consumer confidence and the environment.

\section{Acknowledgements}
We thank Mischa Dohler, Sagar Joglekar, Jung Wing Wan, Dhruv Sengupta and Angus Thompson for proofreading our work at incredibly short notice. We thank Elena Sinel for making us aware that N(eur)IPS was still open for submissions in September.

\Urlmuskip=0mu plus 1mu\relax
\bibliographystyle{unsrtnat}
\bibliography{ref}

\end{document}